\documentclass[11pt]{article}

\usepackage[final]{acl}

\usepackage{times}
\usepackage{latexsym}
\usepackage{amsmath}

\usepackage[T1]{fontenc}

\usepackage[utf8]{inputenc}

\usepackage{microtype}

\usepackage{inconsolata}

\usepackage{graphicx}

\usepackage{booktabs}
\usepackage{graphicx}
\graphicspath{{.}{figures/}} 

\title{Evaluating LLMs for Zeolite Synthesis Event Extraction (ZSEE): A Systematic Analysis of Prompting Strategies}

\author{
  \textbf{Charan Prakash Rathore\textsuperscript{1}}, 
  \textbf{Saumi Ray\textsuperscript{1}},
  \textbf{Dhruv Kumar\textsuperscript{1}}\\ 
  \textsuperscript{1}Birla Institute of Technology and Science, Pilani, India \\
  \small{\textbf{Correspondence:} \href{mailto:dhruv.kumar@pilani.bits-pilani.ac.in}{dhruv.kumar@pilani.bits-pilani.ac.in}}
}

\usepackage{multirow}
\usepackage{booktabs}
\usepackage{graphicx}
\usepackage{adjustbox}   
\usepackage{tabularx}    
\usepackage{placeins}    
\graphicspath{{.}{figures/}}  
\usepackage[font=bf]{caption}

\makeatletter
\newcommand{\safegraphics}[2][\columnwidth]{%
  \IfFileExists{#2}{\includegraphics[width=#1]{#2}}{\fbox{\texttt{#2} missing}}%
}
\makeatother
\usepackage{booktabs}
\usepackage{adjustbox}   
\usepackage{placeins}    
\usepackage{array}       
\newcolumntype{M}[1]{>{\centering\arraybackslash}p{#1}}

\usepackage{enumitem}
\setlist[itemize]{leftmargin=*, itemsep=0pt, topsep=0pt, parsep=0pt, partopsep=0pt}

\usepackage{listings} 
\usepackage{xcolor}   
\usepackage{listings}
\usepackage[most]{tcolorbox}
\tcbuselibrary{listings,skins,breakable}

\definecolor{sand}{RGB}{255,248,225}
\lstdefinestyle{pyfriendly}{
  language=Python,
  basicstyle=\ttfamily\small,
  keywordstyle=\color{blue},
  commentstyle=\color{green!50!black},
  stringstyle=\color{brown},
  showstringspaces=false,
  breaklines=true,
  tabsize=4
}
\newtcblisting{pybubblelistings}{
  breakable, enhanced,
  colback=sand, colframe=sand, boxrule=0pt,
  arc=16pt, outer arc=16pt,
  left=10pt,right=10pt,top=10pt,bottom=10pt,
  listing only,
  listing options={style=pyfriendly}
}

\begin{document}

\maketitle

\begin{abstract}

Extracting structured information from zeolite synthesis experimental procedures is critical for materials discovery, yet existing methods have not systematically evaluated Large Language Models (LLMs) for this domain-specific task. This work addresses a fundamental question: what is the efficacy of different prompting strategies when applying LLMs to scientific information extraction? We focus on four key subtasks: event type classification (identifying synthesis steps), trigger text identification (locating event mentions), argument role extraction (recognizing parameter types), and argument text extraction (extracting parameter values). We evaluate four prompting strategies - zero-shot, few-shot, event-specific, and reflection-based - across six state-of-the-art LLMs (Gemma-3-12b-it, GPT-5-mini, O4-mini, Claude-Haiku-3.5, DeepSeek reasoning and non-reasoning) using the ZSEE dataset of 1,530 annotated sentences. Results demonstrate strong performance on event type classification (80-90\% F1) but modest performance on fine-grained extraction tasks, particularly argument role and argument text extraction (50-65\% F1). GPT-5-mini exhibits extreme prompt sensitivity with 11-79\% F1 variation. Notably, advanced prompting strategies provide minimal improvements over zero-shot approaches, revealing fundamental architectural limitations. Error analysis identifies systematic hallucination, over-generalization, and inability to capture synthesis-specific nuances. Our findings demonstrate that while LLMs achieve high-level understanding, precise extraction of experimental parameters requires domain-adapted models, providing quantitative benchmarks for scientific information extraction.
\end{abstract}

\section{Introduction}

Zeolites are crucial industrial catalysts whose automated synthesis requires extracting structured, machine-readable data from unstructured experimental procedures \cite{jensen2019chemical}. Event extraction - a core information extraction task that identifies specific occurrences or actions mentioned in text along with their associated participants and attributes \cite{ahn2006stages}-offers a systematic approach to structuring procedural knowledge. Argument extraction complements this by identifying and classifying the entities, temporal expressions, and other parameters associated with these events \cite{li2013joint,yang2019exploring}. In the context of zeolite synthesis, event-argument extraction involves identifying synthesis actions (e.g., Add, Stir, Calcine), their textual triggers, and associated arguments such as materials, temperatures, and durations from complex procedural sentences. This task is particularly challenging due to domain-specific terminology, implicit information, complex sentence structures, and the need for precise span identification at the token level.

Traditional approaches to scientific information extraction rely heavily on supervised learning with domain-specific labeled data \cite{luan2018multitask,jain2020scirex}. However, NLP-based information extraction for specialized domains remains limited by scarce annotated datasets and domain-specific complexity. The emergence of Large Language Models (LLMs) pre-trained on vast corpora has demonstrated remarkable capabilities through in-context learning and prompting strategies. This raises a critical question: \textit{can general-purpose LLMs effectively perform specialized scientific information extraction without extensive fine-tuning?} 

Recent work on scientific event extraction has primarily focused on developing specialized neural architectures with domain adaptation. The ZSEE (Zeolite Synthesis Event Extraction) dataset \cite{he-etal-2024-zsee} introduced expert-annotated data for zeolite synthesis procedures and evaluated tailored models like PAIE, while Zeo-Reader \cite{he2025zeoreader} improved representation learning through contrastive learning techniques. Other specialized approaches including AMPERE, DEGREE, and EEQA have demonstrated the effectiveness of carefully designed architectures \cite{du2020event,hsu2022degree,wei2021trigger}. However, a significant gap exists in understanding how general-purpose LLMs perform on such specialized extraction tasks. While ZSEE noted limitations of GPT-3.5-turbo regarding hallucination and over-generalization, no comprehensive systematic evaluation across multiple state-of-the-art LLMs using varied prompting strategies has been conducted.

We present a systematic benchmark study evaluating six contemporary LLMs across four distinct prompting paradigms. Our methodology employs a standardized evaluation framework applied consistently across all models and prompting strategies. We test four prompting approaches: (1) zero-shot prompting with detailed task definitions \cite{kojima2022large}, (2) few-shot prompting with annotated examples \cite{brown2020language}, (3) event-specific prompting with targeted event descriptions, and (4) reflection prompting that encourages self-correction \cite{madaan2023self,shinn2023reflexion}. We evaluate six LLMs: Gemma-3-12b-it \cite{team2024gemma}, GPT-5-mini \cite{openai2024gpt4}, O4-mini \cite{openai2024o1}, Claude-Haiku-3.5 \cite{anthropic2024claude}, DeepSeek-reasoning \cite{deepseek2024r1}, and DeepSeek-non-reasoning \cite{guo2024deepseek}. All models are evaluated on the same 1,530 sentences from the ZSEE dataset, ensuring fair comparison. We measure performance using precision, recall, and F1 scores computed through lemmatization-based subset matching for four subtasks: event type classification, trigger text extraction, argument role identification, and argument text extraction.


Results show LLMs achieve reasonable event type classification (80-90\% F1) but vary significantly on trigger text extraction (60-87\% F1, with GPT-5-mini showing extreme variance: 11-79\%). Argument extraction achieves moderate performance (62-73\% F1 for roles, 57-65\% for texts). Critically, advanced prompting strategies provide only marginal improvements (1-5 percentage points) over zero-shot approaches. Qualitative analysis reveals systematic failures including hallucination, over-generalization, imprecise span boundaries, and confusion between similar argument types, indicating fundamental limitations in precise span-level extraction.

Our work makes the following research contributions:

\begin{itemize}
    \item \textbf{Comprehensive LLM Benchmark:} We provide the first systematic evaluation of multiple state-of-the-art LLMs on scientific event-argument extraction, establishing performance baselines across six models and four prompting strategies.

    \item \textbf{Prompting Strategy Analysis:} We demonstrate empirically that advanced prompting techniques (few-shot, event-specific, reflection) offer minimal improvements over zero-shot prompting for this task, providing important insights for practitioners.

    \item \textbf{Failure Mode Characterization:} Through detailed error analysis, we identify and categorize systematic failure patterns in LLM-based extraction, including hallucination, over-generalization, and span boundary errors.
\end{itemize}

\section{Related Work}

\subsection{Scientific Information Extraction}
Scientific information extraction has evolved from rule-based systems to sophisticated neural approaches. Early work relied on hand-crafted patterns and dictionaries \cite{banko2007open,fader2011identifying}, which achieved high precision but poor recall and required substantial domain expertise. Statistical methods using CRFs and structured SVMs \cite{lafferty2001conditional,finkel2005incorporating} improved generalization but still required extensive feature engineering. Recent neural approaches employ various architectures including BiLSTM-CRF models \cite{lample2016neural}, attention mechanisms \cite{nguyen2016joint}, and transformer-based models \cite{wadden2019entity,lu2020cross}. PAIE introduced prompt-based learning for event extraction, demonstrating that formulating extraction as a QA task can improve performance \cite{ma2022prompt}. DEGREE and AMPERE further advanced structured prediction for event extraction through joint modeling of event types and arguments \cite{hsu2022degree,du2020event}.

Our work differs by evaluating general-purpose LLMs rather than specialized extraction models, focusing on understanding what can be achieved through prompting alone without task-specific fine-tuning. This provides insights into the applicability of foundation models to specialized scientific domains.

\subsection{Domain-Specific Event Extraction}
Domain-specific event extraction has received significant attention in biomedical NLP, where systems extract protein interactions, drug effects, and clinical events \cite{kim2009overview}. ChemDataExtractor and similar systems have demonstrated effective extraction of chemical synthesis information \cite{swain2016chemdataextractor}. The ZSEE dataset introduced a curated benchmark for zeolite synthesis, providing expert annotations and establishing baselines using models like PAIE \cite{ma2022prompt}. Zero-Reader extended this work by incorporating contrastive learning to better handle abstract expressions in experimental descriptions \cite{he2025zeoreader}. These approaches typically require substantial labeled data and domain-specific model architectures.

In contrast, our work explores whether modern LLMs' broad pre-training and few-shot learning capabilities can obviate the need for extensive domain-specific annotation and model engineering. 

\subsection{Large Language Models and Prompting}
Large Language Models have demonstrated remarkable capabilities through scale and pre-training on diverse text \cite{brown2020language,chowdhery2022palm,ouyang2022training}. Research on prompting strategies has shown that careful prompt design can significantly impact performance \cite{liu2023pre,mishra2021reframing}. Few-shot learning enables models to perform tasks from minimal examples \cite{brown2020language}, while chain-of-thought prompting elicits reasoning capabilities \cite{wei2022chain}. Instruction tuning and reinforcement learning from human feedback have further improved instruction following \cite{ouyang2022training,sanh2021multitask}. Self-consistency and self-refinement techniques enable models to improve their outputs iteratively \cite{wang2022self,madaan2023self}.

Despite these advances in prompting techniques, a critical question remains: can prompting strategies alone overcome fundamental limitations in precise information extraction from specialized domains? While prompt engineering has shown promise in improving general task performance across various NLP tasks, the gap between LLM capabilities and specialized model requirements for scientific extraction presents an important area of investigation. Understanding whether advanced prompting techniques like reflection and few-shot learning can bridge this gap for domain-specific extraction tasks motivates our systematic evaluation.

\subsection{Evaluation of LLMs on Structured Tasks}
Recent work has begun evaluating LLMs on structured prediction tasks. Studies have shown mixed results: LLMs excel at tasks with clear contextual signals but struggle with precise span identification and structured output \cite{wei2023chainofthought,zhang2023instruction}. Research on information extraction has found that LLMs can match or exceed supervised models on certain entity recognition tasks but face challenges with relation extraction and event extraction requiring fine-grained understanding \cite{li2023evaluating,wadhwa2023revisiting}. Work on biomedical NLP has similarly found that while LLMs possess domain knowledge, they struggle with precise extraction compared to fine-tuned models \cite{singhal2023large}.

Our work contributes to this literature by providing a comprehensive evaluation specifically on scientific procedural text, an under-studied domain. Unlike prior work focusing on single models or tasks, we systematically compare multiple models across multiple prompting strategies, providing clearer insights into the performance ceiling achievable with current LLM technology on specialized extraction tasks.

\section{Methodology}

\begin{figure*}[t] 
    \centering
    \includegraphics[width=\textwidth]{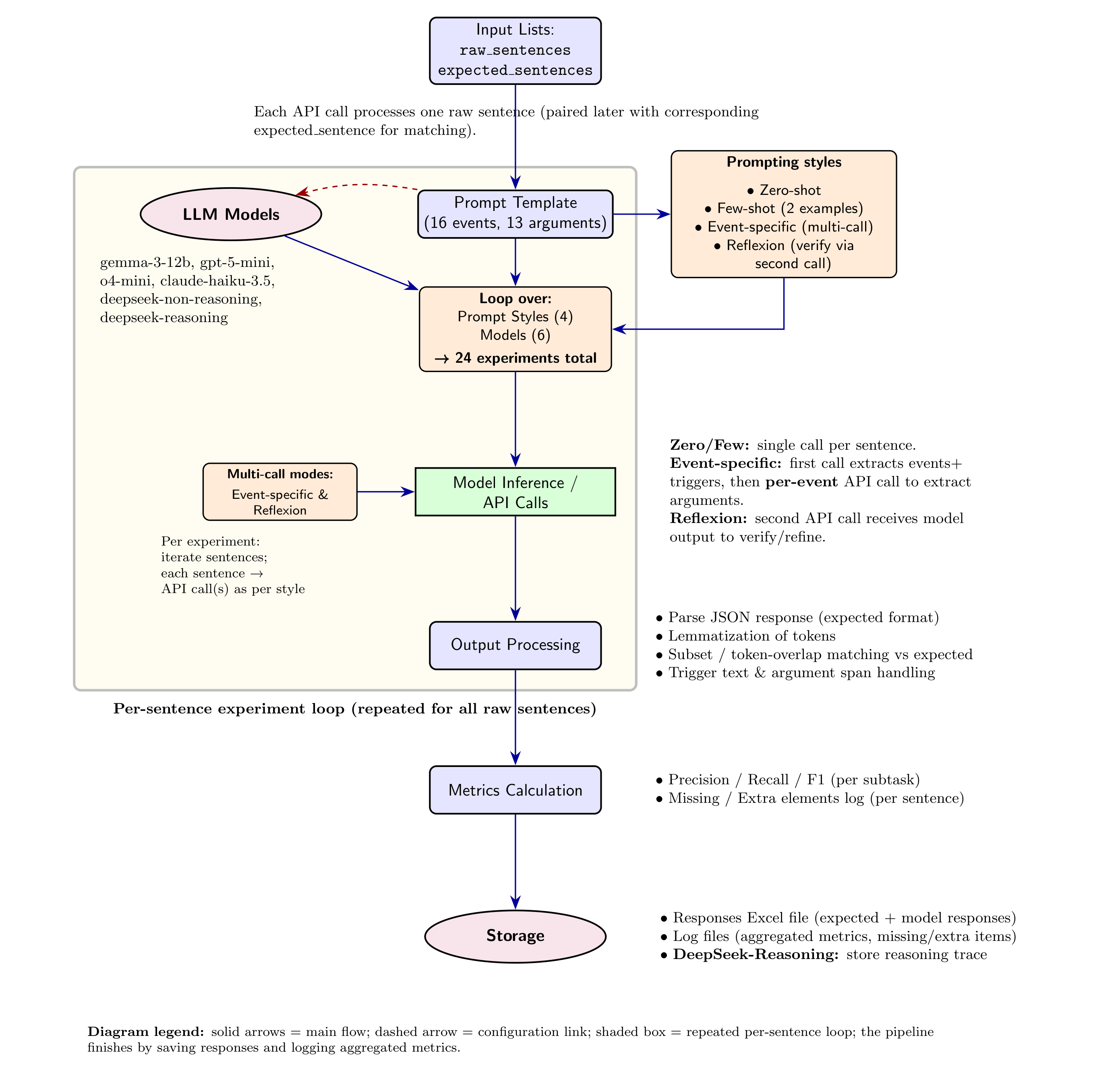}
    \caption{\textbf{Pipeline for event and argument extraction using LLMs on the ZSEE dataset.}}
    \label{fig:pipeline}
\end{figure*}

Our methodology focuses on systematic evaluation of LLM-based extraction using standardized prompting strategies. An overview of the complete experimental pipeline is illustrated in \textbf{Figure~\ref{fig:pipeline}}. We describe the task formulation, prompting approaches, and evaluation framework.

\subsection{Task Formulation}
Given a scientific procedural sentence, our task is to extract structured information in JSON format consisting of events and their arguments. Each event comprises: \textbf{(1)} Event Type: The action category (e.g., Add, Stir, Calcine). \textbf{(2)} Trigger Text: The exact word(s) in the sentence indicating the event. \textbf{(3)} Arguments: A list of (role, text) pairs where role specifies the argument type (e.g., material, temperature) and text provides the exact span from the sentence.

\noindent For example, given the sentence:
\begin{quote}
\textit{``The calcined samples (0.3 g) were dispersed in the ammonium nitrate solution (100 mL) and then stirred at 500 rpm and room temperature.''}
\end{quote}

The expected output contains two events:
\begin{pybubblelistings}
{
  "events": [
    {
      "event_type": "Add",
      "trigger_text": "dispersed",
      "arguments": [
        {"role": "material", 
         "text": "calcined samples (0.3 g)"},
        {"role": "material", 
         "text": "ammonium nitrate solution
                 (100 mL)"}
      ]
    },
    {
      "event_type": "Stir",
      "trigger_text": "stirred",
      "arguments": [
        {"role": "revolution", 
         "text": "500 rpm"},
        {"role": "temperature", 
         "text": "room temperature"}
      ]
    }
  ]
}
\end{pybubblelistings}

\subsection{Prompting Strategies}

We evaluate four prompting approaches of increasing complexity:

\subsubsection{Zero-Shot Prompting}
The zero-shot prompt\cite{kojima2022large} provides complete task definition including \textbf{(1)} JSON schema specification, \textbf{(2)} all 16 possible event types with detailed definitions, \textbf{(3)} all 13 argument roles with descriptions, and \textbf{(4)} output format constraints (JSON only, no explanations). The prompt emphasizes extracting only explicitly stated information from the current sentence, avoiding inference from context. This baseline approach tests the model's ability to perform the task from instructions alone.

\subsubsection{Few-Shot Prompting}
Building on zero-shot prompting, we augment the prompt with 2-3 annotated examples demonstrating correct extraction\cite{brown2020language}. Examples are selected to cover diverse event types and sentence structures. Each example includes the input sentence and expected JSON output, providing the model with concrete templates for the extraction format.

\subsubsection{Event-Specific Prompting}
This approach performs extraction in two conditional stages. \textbf{(1)} The first API call extracts only event types and their trigger texts. \textbf{(2)} For each detected event, a separate follow-up call is issued, using an event-specific prompt containing detailed argument role definitions and extraction instructions. This enables focused argument extraction tailored to each event rather than using a single generic prompt.


\subsubsection{Reflection Prompting}
Also known as self-correction or self-refinement prompting\cite{madaan2023self,shinn2023reflexion}, this technique involves a two-stage process: \textbf{(1)} generating an initial extraction using the zero-shot prompt, and \textbf{(2)} providing the initial output back to the model with a prompt requesting verification and correction of potential errors.

The reflection prompt asks the model to check for hallucination, verify trigger text accuracy, ensure argument roles match the event type, and validate that extracted text appears verbatim in the input sentence.

\subsection{Models Evaluated}

We evaluate six contemporary LLMs representing different architectural approaches and scales: (1) \textbf{Gemma-3-12b-it} \cite{team2024gemma}, Google's 12B parameter instruction-tuned model; (2)  \textbf{GPT-5-mini} \cite{openai2024gpt4}, an efficient GPT-5 variant from OpenAI; (3) \textbf{O4-mini} \cite{openai2024o1}, a reasoning-optimized compact model; (4) \textbf{Claude-Haiku-3.5} \cite{anthropic2024claude}, Anthropic's fast and cost-efficient model; (5) \textbf{DeepSeek-Reasoning} \cite{deepseek2024r1}, designed with enhanced reasoning capabilities; and (6) \textbf{DeepSeek-Non-Reasoning} \cite{guo2024deepseek}, used as a standard DeepSeek baseline for comparison. All models are accessed via their respective APIs using default temperature settings (0.7 for most models) to balance determinism with natural language generation.

\subsection{Evaluation Metrics}

We evaluate extraction quality using precision, recall, and F1 score computed through lemmatization-based subset matching. This approach accounts for morphological variations while ensuring that extracted text substantively matches the ground truth.

\textbf{Metric Calculation :}
For each subtask (event types, trigger texts, argument roles, argument texts), we \textbf{(1)} lemmatize all expected and predicted values, \textbf{(2)} count the frequency of each unique lemmatized value, and \textbf{(3)} compute correct predictions as the minimum of expected and predicted counts for each value.






\section{Experimental Setup}

\subsection{Dataset}
We use the ZSEE (Zeolite Synthesis Event Extraction) dataset \cite{he-etal-2024-zsee}, which contains 1,530 sentences from zeolite synthesis procedures extracted from scientific publications. Each sentence is annotated by domain experts with:

\begin{itemize}
\item \textbf{16 Event Types:} Add, Stir, Wash, Dry, Calcine, Crystallize, Particle Recovery, Heat, Set pH, Rotate, Sonicate, Seal, Transfer, Age, Cool, React
\item \textbf{13 Argument Types:} material, temperature, duration, container, sample, solvent, condition, revolution, times, pH, rate, pressure, revolution\_text
\end{itemize}

The dataset is publicly available at \url{https://github.com/Hi-0317/ZSEE} and represents a challenging test case due to domain-specific terminology, complex sentence structures, and implicit information requiring domain knowledge.

\subsection{Baselines}
We compare our LLM-based approaches against the state-of-the-art results reported in the ZSEE paper:

\begin{itemize}
\item \textbf{PAIE} (Prompt-based Argument Interaction for Event Extraction): Achieved 92\% F1 on event extraction and 74\% F1 on argument extraction
\item \textbf{Zero-Reader}: Further improved performance through contrastive learning, particularly for argument extraction
\end{itemize}

These specialized models represent the current performance ceiling for this task and serve as the standard against which LLM performance should be evaluated.

\subsection{Implementation Details}
Our experimental design systematically evaluates each combination of model and prompting strategy:
\begin{itemize}
\item \textbf{Models}: 6 (Gemma-3-12b-it, GPT-5-mini, O4-mini, Claude-Haiku-3.5, DeepSeek-Reasoning, DeepSeek-Non-Reasoning)
\item \textbf{Prompting Strategies}: 4 (Zero-shot, Few-shot, Event-specific, Reflection)


\item \textbf{Samples per Condition}: 1,530 sentences. For each condition, we \textbf{(1)} process sentences sequentially via API calls, \textbf{(2)} parse JSON responses and extract predictions for each subtask, \textbf{(3)} compare predictions against gold annotations using lemmatization-based matching, \textbf{(4)} aggregate metrics across all sentences, and \textbf{(5)} store detailed error information including missing and extra predictions.
\end{itemize}

All experiments were conducted using Python 3.9 with API calls to respective model providers. We enforced structured output by explicitly requesting JSON format and implemented basic parsing validation. In cases where models produced malformed JSON, we attempted automatic correction; sentences with unparseable output were marked as failures and contributed zero to all metrics. For DeepSeek-Reasoning specifically, we also captured the model's internal reasoning traces to enable analysis of the model's decision-making process.


\section{Results}

\begin{table*}[t]
\centering
\caption{F1 Scores (\%) for All Models and Prompting Strategies}
\label{tab:all_results}
\small
\begin{tabular}{llcccccccc}
\toprule
& & \multicolumn{4}{c}{\textbf{Event Type}} & \multicolumn{4}{c}{\textbf{Trigger Text}} \\
\cmidrule(lr){3-6} \cmidrule(lr){7-10}
\textbf{Model} & \textbf{Prompt} & \textbf{Zero} & \textbf{Few} & \textbf{Event} & \textbf{Refl} & \textbf{Zero} & \textbf{Few} & \textbf{Event} & \textbf{Refl} \\
\midrule
Gemma-3-12b-it & & 86.52 & 85.38 & 80.7 & 84.81 & 70.4 & 87.56 & 79.45 & 72.04 \\
GPT-5-mini & & 88.55 & 86.51 & 86.06 & 88.75 & 24.59 & 79.75 & 11.51 & 29.19 \\
O4-mini & & 87.04 & 85.45 & 86.99 & 87.86 & 87.04 & 85.45 & 74.06 & 76.46 \\
Claude-Haiku-3.5 & & 88.7 & 87.46 & 83.55 & 86.46 & 61.74 & 84.52 & 80.28 & 62.34 \\
DeepSeek-NR & & 80.56 & 79.34 & 84.07 & 84.42 & 71.82 & 82.84 & 80.36 & 74.63 \\
DeepSeek-R & & 80.86 & 79.6 & 78.7 & 79.01 & 77.15 & 83.48 & 80.85 & 76.07 \\
\midrule
& & \multicolumn{4}{c}{\textbf{Argument Roles}} & \multicolumn{4}{c}{\textbf{Argument Texts}} \\
\cmidrule(lr){3-6} \cmidrule(lr){7-10}
Gemma-3-12b-it & & 64.28 & 74.65 & 68.65 & 67.72 & 56.54 & 65.94 & 56.39 & 59.58 \\
GPT-5-mini & & 62.63 & 83.92 & 66.11 & 69.19 & 66.72 & 58.5 & 57.24 & 62.47 \\
O4-mini & & 68.46 & 70.37 & 69.68 & 71.4 & 62.72 & 64.72 & 61.78 & 65.19 \\
Claude-Haiku-3.5 & & 68.09 & 71.1 & 65.47 & 67.3 & 58.59 & 64.79 & 56.49 & 59.4 \\
DeepSeek-NR & & 62.96 & 68.69 & 67.51 & 63.29 & 56.44 & 61.03 & 58.08 & 57.69 \\
DeepSeek-R & & 70.06 & 70.9 & 69.26 & 72.01 & 60.72 & 59.81 & 54.79 & 61.9 \\
\bottomrule

\end{tabular}
\end{table*}

We present comprehensive results across all models and prompting strategies, organized by subtask. Table~\ref{tab:all_results} shows F1 scores as the primary metric, with precision and recall provided in Appendix A.

\textbf{Event Type Extraction: }As shown in Table~\ref{tab:all_results}, event type extraction represents the highest-performing subtask, with all models achieving F1 scores between 80\% and 90\%. This suggests that identifying action categories from contextual cues is within current LLM capabilities. However, prompting strategy shows minimal impact: the difference between best and worst prompts averages only 2-3 percentage points per model.

Surprisingly, zero-shot prompting often outperforms more complex strategies. This may indicate that detailed event definitions or examples introduce confusion rather than clarity, or that models already possess sufficient prior knowledge about experimental actions.

\textbf{Trigger Text Extraction:} The results in Table~\ref{tab:all_results} show that trigger text extraction exhibits the most significant variation across both models and prompting strategies. Few-shot prompting substantially outperforms other approaches, averaging 83.93\% F1 compared to 65-68\% for other strategies. This suggests that concrete examples of valid trigger words help models identify the correct spans.

GPT-5-mini exhibits extreme sensitivity to prompting strategy, with F1 scores ranging from 11.51\% (event-specific) to 79.75\% (few-shot). In contrast, the other five models demonstrate more stable performance across prompting strategies, consistently achieving 60-87\% F1 regardless of prompt design.
This anomalous behavior of GPT-5-mini suggests that it struggles to identify trigger text spans from task descriptions alone but can effectively learn from concrete examples. 

Despite few-shot improvements, models still make systematic errors including: (1) extracting related but incorrect words (e.g., ``placed'' instead of ``dispersed''), (2) extracting multiple words when a single word is correct, and (3) omitting trigger words entirely when event type is identified correctly. These errors indicate difficulty with precise span boundary identification even when the general concept is understood.

\textbf{Argument Role Extraction:} As presented in Table~\ref{tab:all_results}, argument role identification achieves moderate performance (66-73\% F1), with few-shot prompting again providing the largest benefit. Common errors include: (1) confusing similar roles (e.g., ``material'' vs. ``sample'', ``revolution'' vs. ``revolution\_text''), (2) hallucinating arguments not present in the sentence, and (3) missing implicit arguments that humans recognize from domain knowledge.

\textbf{Argument Text Extraction:} As reflected in Table~\ref{tab:all_results}, argument text extraction shows the most consistent (but modest) performance across strategies, with F1 scores clustered around 57-62\%. This represents the most challenging subtask as it requires identifying exact spans of potentially complex phrases. Models struggle with: (1) boundary errors (e.g., including or excluding parenthetical units), (2) partial extraction (extracting part of a compound phrase), and (3) paraphrasing (restating information rather than extracting verbatim text).

\textbf{Impact of Prompting Strategies:}
Few-shot prompting provides the most substantial benefit over the zero-shot baseline, particularly for trigger text extraction, improving performance by approximately +18\% F1. Argument role and argument text extraction also show moderate gains under few-shot prompting (+7\% and +2\% F1, respectively). In contrast, event-specific prompting offers negligible improvements and even reduces performance on two subtasks, while reflection prompting yields only marginal gains. Overall, these results indicate that performance is primarily constrained by model capability rather than prompt engineering: concrete examples help with token-level span decisions, but elaborate task descriptions or self-correction mechanisms do not substantially improve quality.

\section{Discussion}

Our results reveal several important insights about LLM capabilities and limitations for scientific information extraction.

\textbf{The Abstraction Gap:} LLMs excel at abstract classification (event types) but struggle with concrete extraction (trigger texts, argument spans). This suggests a fundamental abstraction gap: models learn high-level semantic patterns effectively but have difficulty grounding these patterns in precise textual spans. This gap persists even with few-shot examples providing explicit templates.

\textbf{1. Hallucination: Inventing Non-Existent Events -}
Models frequently extract events not defined in our schema, creating event types based on surface-level textual cues rather than adhering to the specified taxonomy. This represents a form of hallucination where models invent categories not present in the task definition.

\textbf{Example:} Given the sentence \textit{``The calcined samples (0.3 g) were dispersed in the ammonium nitrate solution (100 mL)...''},  the model extracted event type ``Disperse'' instead of the correct ``Add''. While ``dispersed'' is indeed the trigger text for an Add event in our schema, the model hallucinated a new event type not among the 16 defined categories.

Errors like this suggest models rely on lexical matching (e.g., seeing ``mixed'' → creating ``Mix'' event) rather than understanding the semantic mapping between trigger words and event categories.

\textbf{2. Over-Generalization: Extracting Implicit Information -}
Models frequently extract implicit information that human annotators deliberately excluded. This reflects a tendency to ``fill in gaps'' using prior knowledge about typical experimental procedures rather than strictly extracting explicitly stated information.

\textbf{Example:} In \textit{``The calcined samples (0.3 g) were dispersed in the ammonium nitrate solution (100 mL) and then stirred at 500 rpm...''},  the model extracted ``then'' as a duration argument. However, ``then'' is merely a temporal connector indicating sequence, not an explicit duration specification. The model over-generalized from the sequential nature of the text.

 Such errors indicate that while domain experts followed strict annotation guidelines to capture only explicit information, LLMs incorporate inferential reasoning that goes beyond what is textually stated.

\textbf{3. Imprecise Span Boundaries -}
Models consistently struggle to identify precise span boundaries for trigger texts, often extracting entire clauses or phrases instead of the specific action words. This represents one of the most pervasive failure patterns across all models.

\textbf{Example:} Given \textit{``25 g of sodium silicate (26.5\% SiO2) was added to 60 g water and stirred for 15 min...''},  the expected trigger texts are simply [``added'', ``stirred'']. However, the model extracted extended spans: [``25 g of sodium silicate (26.5\% SiO2) was added to 60 g water'', ``stirred for 15 min'']. The model incorporated surrounding context rather than isolating the precise action verbs.

This pattern suggests models lack the fine-grained token-level precision needed for span identification, instead treating extraction as a clause or phrase-level task.

\textbf{4. Confusion Between Similar Argument Types -}
Models frequently confuse semantically related argument roles, particularly when categories have overlapping characteristics or could be contextually ambiguous.

\textbf{Example:} In \textit{``The calcined samples (0.3 g) were dispersed in the ammonium nitrate solution (100 mL) and then stirred at 500 rpm...''},  the expected argument roles are [``material'', ``material'', ``revolution'', ``temperature'']. However, the model classified the second material (``ammonium nitrate solution'') as ``solvent'' and misidentified other arguments. This confusion between ``material'' and ``solvent'' reflects the semantic proximity of these categories in experimental contexts.

The example illustrates that while models understand general experimental semantics, they struggle with the fine-grained distinctions necessary for precise role classification in domain-specific schemas.

\textbf{Limited Impact of Prompting:}
Our most surprising finding is the limited impact of advanced prompting strategies. Few-shot prompting provides moderate benefits for span-based tasks, but event-specific definitions and reflection prompting offer minimal improvements. This suggests that models already possess substantial domain knowledge from pre-training, but the challenge lies in executing precise extraction rather than understanding task requirements.


\textbf{Implications for Scientific Text Mining:}
These findings have important practical implications. First, practitioners should not expect dramatic improvements from prompt engineering alone; fundamental model limitations constrain performance. Second, hybrid approaches combining LLM outputs with post-processing or structured extraction methods may be more effective than pure LLM-based pipelines. Third, specialized models fine-tuned on domain-specific data remain necessary for applications requiring high precision.

\textbf{Comparison with Specialized Models:}
Comparing our results to the ZSEE baselines reveals a substantial performance gap. PAIE achieved 92\% F1 on event extraction compared to our 83-85\% average, and 74\% on argument extraction compared to our 57-73\% depending on subtask. This 8-17 percentage point gap represents a significant practical difference and demonstrates that specialized architectures with domain adaptation still substantially outperform general-purpose LLMs on precise extraction tasks.

\section{Conclusion and Future Work}

This study provides the first systematic evaluation of six LLMs and four prompting strategies for ZSEE event-argument extraction. Models perform well on event classification but struggle with precise span extraction, and advanced prompting offers minimal advantage over zero-shot. GPT-5-mini shows extreme prompt sensitivity, and common failures include hallucination, over-generalization, and boundary errors. A substantial performance gap remains between general-purpose LLMs and specialized extraction models. Future work will explore hybrid pipelines, lightweight domain adaptation, and evaluation across additional scientific domains.

\section{Limitations}

Our study has several limitations that should be considered when interpreting results:

\textbf{Dataset Scope:} We evaluate on a single dataset (ZSEE) from one domain (zeolite synthesis). While this provides a controlled evaluation, generalization to other scientific domains remains unclear.

\textbf{Prompting Strategies:} We evaluate four prompting approaches, but the space of possible prompt designs is vast. More sophisticated techniques like chain-of-thought prompting with intermediate reasoning steps or multi-step refinement with separate extraction and validation phases might yield better results.

\textbf{Model Configuration:} We use default API parameters (temperature, top-p) for all models. Different sampling strategies or temperature settings might affect results, though preliminary experiments suggested minimal impact.

\textbf{Computational Constraints:} Full evaluation across 24 conditions required substantial API costs. This constrained our ability to explore additional models or more sophisticated prompting strategies involving multiple inference passes.

\section{Acknowledgements}
We thank the creators of the ZSEE dataset for making their data publicly available. We are also grateful for access to API platforms that enabled this evaluation. This work was conducted as part of academic research into applications of large language models for scientific information extraction.

The authors also wish to acknowledge the use of ChatGPT and Claude in improving the presentation and grammar of the paper. The paper remains an accurate representation of the authors’ underlying contributions.


\bibliography{references}

\section{Appendix}
\appendix

\section{Complete Results Tables}

Tables \ref{tab:complete_precision} and \ref{tab:complete_recall} present complete precision and recall values for all models, prompting strategies, and subtasks.

\begin{table*}[h]
\centering
\caption{Complete Precision Results (\%) for All Models and Prompting Strategies}
\label{tab:complete_precision}
\small
\begin{tabular}{llcccccccc}
\toprule
& & \multicolumn{4}{c}{\textbf{Event Type}} & \multicolumn{4}{c}{\textbf{Trigger Text}} \\
\cmidrule(lr){3-6} \cmidrule(lr){7-10}
\textbf{Model} & \textbf{Prompt} & \textbf{Zero} & \textbf{Few} & \textbf{Event} & \textbf{Refl} & \textbf{Zero} & \textbf{Few} & \textbf{Event} & \textbf{Refl} \\
\midrule
Gemma-3-12b-it & & 87.74 & 84.71 & 81.33 & 85.51 & 71.67 & 88.59 & 80.05 & 72.82 \\
GPT-5-mini & & 88.86 & 85.42 & 83.85 & 88.69 & 24.68 & 78.69 & 11.25 & 29.24 \\
O4-mini & & 88.82 & 85.99 & 87.1 & 88.77 & 75.04 & 84.35 & 73.92 & 76.92 \\
Claude-Haiku-3.5 & & 90.05 & 88.11 & 86.46 & 87.86 & 62.6 & 85.01 & 83.02 & 63.21 \\
DeepSeek-NR & & 84.01 & 82.29 & 86.45 & 87.29 & 74.67 & 85.12 & 82.62 & 76.93 \\
DeepSeek-R & & 83.65 & 83.17 & 81.16 & 81.85 & 79.7 & 86.99 & 83.09 & 78.47 \\
\midrule
& & \multicolumn{4}{c}{\textbf{Argument Roles}} & \multicolumn{4}{c}{\textbf{Argument Texts}} \\
\cmidrule(lr){3-6} \cmidrule(lr){7-10}
Gemma-3-12b-it & & 56.79 & 73.12 & 70.36 & 59.95 & 50.16 & 64.48 & 57.25 & 52.93 \\
GPT-5-mini & & 52.5 & 56.31 & 58.65 & 61.23 & 47.61 & 49.82 & 50.62 & 55.3 \\
O4-mini & & 60.49 & 65.71 & 64.08 & 64.67 & 55.74 & 58.87 & 56.69 & 59.18 \\
Claude-Haiku-3.5 & & 60.25 & 64.61 & 62.16 & 59.41 & 52.59 & 59.16 & 53.4 & 52.61 \\
DeepSeek-NR & & 55.83 & 65.25 & 67.11 & 56.8 & 50 & 57.37 & 56.59 & 51.71 \\
DeepSeek-R & & 65.95 & 71.95 & 69.52 & 69.37 & 56.58 & 58.98 & 53.33 & 58.97 \\
\bottomrule
\end{tabular}
\end{table*}

\begin{table*}[h]
\centering
\caption{Complete Recall Results (\%) for All Models and Prompting Strategies}
\label{tab:complete_recall}
\small
\begin{tabular}{llcccccccc}
\toprule
& & \multicolumn{4}{c}{\textbf{Event Type}} & \multicolumn{4}{c}{\textbf{Trigger Text}} \\
\cmidrule(lr){3-6} \cmidrule(lr){7-10}
\textbf{Model} & \textbf{Prompt} & \textbf{Zero} & \textbf{Few} & \textbf{Event} & \textbf{Refl} & \textbf{Zero} & \textbf{Few} & \textbf{Event} & \textbf{Refl} \\
\midrule
Gemma-3-12b-it & & 88.37 & 88.59 & 82.55 & 87.32 & 71.54 & 90.63 & 81.21 & 73.89 \\
GPT-5-mini & & 90.36 & 90.56 & 91.69 & 90.97 & 25.01 & 85.96 & 12.18 & 29.57 \\
O4-mini & & 87.35 & 85.45 & 88.81 & 88.91 & 73.85 & 85.68 & 75.66 & 76.98 \\
Claude-Haiku-3.5 & & 89.39 & 89.06 & 84.08 & 87.2 & 62.19 & 86.16 & 80.74 & 62.86 \\
DeepSeek-NR & & 79.76 & 79.02 & 84.73 & 84.16 & 71.21 & 81.98 & 81.02 & 74.51 \\
DeepSeek-R & & 80.08 & 78.44 & 78.8 & 78.33 & 76.41 & 82.42 & 81.11 & 75.57 \\
\midrule
& & \multicolumn{4}{c}{\textbf{Argument Roles}} & \multicolumn{4}{c}{\textbf{Argument Texts}} \\
\cmidrule(lr){3-6} \cmidrule(lr){7-10}
Gemma-3-12b-it & & 82.54 & 81.78 & 73.44 & 85 & 71.37 & 71.82 & 60.7 & 73.92 \\
GPT-5-mini & & 85.22 & 85 & 84.48 & 85.81 & 76.38 & 77 & 72.94 & 77 \\
O4-mini & & 85.01 & 84.42 & 82.84 & 85.64 & 76.86 & 76.75 & 73.26 & 77.35 \\
Claude-Haiku-3.5 & & 84.67 & 84.53 & 78.28 & 84.23 & 71.7 & 76.1 & 67.19 & 73.15 \\
DeepSeek-NR & & 79.03 & 79.12 & 76.85 & 78.32 & 70.65 & 70.74 & 66.39 & 71 \\
DeepSeek-R & & 81.48 & 82.44 & 81.11 & 75.57 & 70.66 & 65.97 & 62.3 & 70.31 \\
\bottomrule
\end{tabular}
\end{table*}

\section{Prompt Templates}

This appendix presents the prompt templates used for each of the four prompting strategies evaluated in our experiments.

\subsection*{B.1 Zero-Shot Prompt Template}

\begin{pybubblelistings}
You are an expert assistant that converts scientific procedure 
sentences into structured JSON. Follow this schema exactly:
{
  "events": [
    {
      "event_type": "...",
      "trigger_text": "...",
      "arguments": [
        {"role": "...", "text": "..."},
        {"role": "...", "text": "..."}
      ]
    }
  ]
}

Event types: Add, Stir, Wash, Dry, Calcine, Crystallize, 
Particle Recovery, Heat, Set pH, Rotate, Sonicate, Seal, 
Transfer, Age, Cool, React.

Roles: material, temperature, duration, container, sample, 
solvent, condition, revolution, times, PH, rate, pressure, 
revolution_text.

You need to extract only those events that are happening in 
the same sentence, not the ones being carried forward from 
previous ones.

[Event Definitions:]
Add: materials are added to the container at a specific 
temperature (arguments: material, temperature, container)

Stir: mixture is stirred with full contact (arguments: 
duration, temperature, revolution, sample)

Age: waiting a period of time for the reaction (arguments: 
duration, temperature, revolution, pressure)

Wash: product is washed several times with some solvent 
(arguments: solvent, times, sample)

Dry: product is dried in the container (arguments: duration, 
temperature, container, condition)

Calcine: product is calcined at high temperature (arguments: 
duration, temperature, container, sample, condition)

Particle Recovery: filtration operations to recover clean 
product (arguments: material, duration, revolution)

Set PH: product is brought to a specific pH value (arguments: 
material, PH)

Cool: temperature is reduced to a specific value (arguments: 
duration, temperature, container, sample, condition)

Heat: temperature is increased to a specific value (arguments: 
duration, temperature, container, sample, pressure, revolution, 
rate)

Crystallize: amorphous compound is converted to crystalline 
state (arguments: duration, temperature, container, pressure, 
revolution)

Transfer: product is transferred from one container to another 
(arguments: sample, container)

Seal: product is kept in a sealed container (arguments: sample, 
container)

Sonicate: product is washed by ultrasound (arguments: sample, 
solvent)

React: ordinary reactions not specifically described in zeolite 
synthesis corpus (arguments: duration, temperature, material, 
condition)

Rotate: direct rotation of a container (arguments: duration, 
temperature, container, revolution)

[Argument Role Definitions:]
duration, temperature, pressure: indicate duration, temperature 
and pressure of the experiment

material: compounds, both liquid and solid, added during 
operations

container: container where synthesis action is carried out

sample: subject of the reaction, different from material

solvent: solvent to which the washing product is added

times: number of washings

condition: specific conditions under which reaction is operated

revolution: specific revolution per minute value

revolution_text: abstract textual representation of rotation

rate: temperature increase rate to a specific value

PH: specific pH value of the product

Input sentence: {sentence}
Output only the JSON but in dictionary format. Do not add 
explanations.
\end{pybubblelistings}

\subsection*{B.2 Few-Shot Prompt Template}

The few-shot prompt uses the same base template as zero-shot, but includes two examples before the input sentence:

\begin{pybubblelistings}
[Same base template as zero-shot, followed by:]

Here are two examples of the required extraction:

Sentence-1: A solution of 24.5 g (0.173 mol) of methyl iodide 
was added and the reaction mixture was stirred at room 
temperature for three days, then new excess of methyl iodide 
(0.173 mol) was added and stirred at room temperature for 
3 days.

Output-1:
{
  "events": [
    {
      "event_type": "Add",
      "trigger_text": "added",
      "arguments": [
        {"role": "material", "text": "solution of 24.5 g 
         ( 0.173 mol ) of methyl iodide"}
      ]
    },
    {
      "event_type": "Add",
      "trigger_text": "added",
      "arguments": [
        {"role": "material", "text": "methyl iodide 
         ( 0.173 mol )"}
      ]
    },
    {
      "event_type": "Stir",
      "trigger_text": "stirred",
      "arguments": [
        {"role": "sample", "text": "mixture"},
        {"role": "duration", "text": "three days"},
        {"role": "temperature", "text": "room temperature"}
      ]
    },
    {
      "event_type": "Stir",
      "trigger_text": "stirred",
      "arguments": [
        {"role": "duration", "text": "3 days"},
        {"role": "temperature", "text": "room temperature"}
      ]
    }
  ]
}

Sentence-2: After this time, the autoclave was cooled down, 
and the mixture was filtered, washed with water and dried 
at 100 C.

Output-2:
{
  "events": [
    {
      "event_type": "Wash",
      "trigger_text": "washed",
      "arguments": [
        {"role": "solvent", "text": "water"}
      ]
    },
    {
      "event_type": "Dry",
      "trigger_text": "dried",
      "arguments": [
        {"role": "temperature", "text": "100 C"}
      ]
    },
    {
      "event_type": "Particle Recovery",
      "trigger_text": "filtered",
      "arguments": [
        {"role": "material", "text": "mixture"}
      ]
    },
    {
      "event_type": "Cool",
      "trigger_text": "cooled",
      "arguments": []
    }
  ]
}

Input sentence: {sentence}
Output only the JSON but in dictionary format. Do not add 
explanations.
\end{pybubblelistings}

\subsection*{B.3 Event-Specific Prompt Template}

This approach uses a two-stage extraction process. First, events and triggers are identified:

\begin{pybubblelistings}
[Stage 1 - Event Identification:]

You are an expert assistant that identifies events in 
scientific procedure sentences. Extract only the event types 
and trigger texts from the sentence that are happening in that 
very same sentence.

For example in the sentence: "The calcined samples (0.3 g) 
were dispersed in the ammonium nitrate solution (100 mL) and 
then stirred at 500 rpm and room temperature."
The word "calcined" is not a calcine event here because it 
isn't actually happening in the same sentence. It is in 
context of other previous process.

Follow this schema exactly:
{
  "events": [
    {
      "event_type": "...",
      "trigger_text": "..."
    }
  ]
}

Event types: Add, Stir, Wash, Dry, Calcine, Crystallize, 
Particle Recovery, Heat, Set pH, Rotate, Sonicate, Seal, 
Transfer, Age, Cool, React.

[Brief event definitions - same as zero-shot but without 
argument specifications]

Input sentence: {sentence}
Output only the JSON format with event_type and trigger_text. 
Do not add explanations.

[Stage 2 - Argument Extraction (per event):]

You are an expert assistant that extracts argument roles and 
texts for the "{event_type}" event in scientific procedure 
sentences.

For the {event_type} event with trigger text "{trigger_text}" 
in the sentence, extract ONLY the following arguments if they 
are present:
[List of valid arguments for this specific event type]

Follow this schema exactly:
{
  "arguments": [
    {"role": "...", "text": "..."},
    {"role": "...", "text": "..."}
  ]
}

IMPORTANT: Extract ONLY the arguments that are actually 
mentioned in the sentence. Do not extract arguments that are 
not present. Only use the roles: [specific roles for this 
event].

Example:
Sentence: For the synthesis of SAPO, fumed silica (Aerosil, 
Degussa) was added to the aqueous solution, and the mixture 
was stirred for 2 h.
Expected output: {"event_type": "Add", "trigger_text": "added", 
"arguments": [{"role": "material", "text": "fumed silica 
( Aerosil , Degussa )"}]}

Sentence: {sentence}
Trigger Text: {trigger_text}

Output only the JSON format. Do not add explanations.
\end{pybubblelistings}

\textit{Note: Stage 2 is repeated for each identified event, with event-specific argument lists provided for each of the 16 event types.}

\subsection*{B.4 Reflexion Prompt Template}

The reflexion approach uses a two-pass system. The first pass uses the zero-shot prompt (Section A.1), followed by a verification pass:

\begin{pybubblelistings}
[Pass 1: Same as Zero-Shot prompt in B.1]

[Pass 2 - Verification and Refinement:]

You are an expert reviewer that checks and corrects event 
extraction from scientific procedure sentences.

You will be given:
1. The original sentence
2. An initial event extraction attempt

Your task is to carefully review the initial extraction and 
provide a corrected version that follows the exact same JSON 
schema.

CRITICAL REQUIREMENTS:
1. Extract ONLY events that are happening in the given sentence, 
   not events mentioned as past context
2. Each event must have the correct event_type and trigger_text
3. Include ALL relevant arguments for each event
4. Use only the specified argument roles for each event type
5. Ensure argument texts are extracted exactly as they appear

Event Type Definitions and Their Valid Arguments:

Add: material, temperature, container
Stir: duration, temperature, revolution, sample
Wash: solvent, times, sample
Dry: duration, temperature, container, condition
Calcine: duration, temperature, container, sample, condition
Crystallize: duration, temperature, container, pressure, 
revolution
Particle Recovery: material, duration, revolution
Heat: duration, temperature, container, sample, pressure, 
revolution, rate
Set pH: material, PH
Rotate: duration, temperature, container, revolution
Sonicate: sample, solvent
Seal: sample, container
Transfer: sample, container
Age: duration, temperature, revolution, pressure
Cool: duration, temperature, container, sample, condition
React: duration, temperature, material, condition

Argument Role Definitions:
- duration: how long the process lasts
- temperature: temperature value with units
- pressure: pressure value with units
- material: compounds being added or used
- container: vessel where action occurs
- sample: what is being processed (different from material)
- solvent: liquid used for washing/cleaning
- times: number of repetitions
- condition: specific conditions (e.g., "in air", "under vacuum")
- revolution: rotation speed (RPM)
- rate: rate of change (e.g., heating rate)
- PH: pH value

REVIEW CHECKLIST:
1. Are there any missed events in the sentence?
2. Are all identified events actually happening in this sentence?
3. Are event types correctly identified?
4. Are trigger texts accurate?
5. Are all possible arguments extracted for each event?
6. Are argument roles correct according to the definitions?
7. Are argument texts extracted exactly as written?

Original sentence: {sentence}

Initial extraction:
{initial_json_str}

Please provide the corrected extraction in the exact same JSON 
format. If the initial extraction was perfect, return it 
unchanged. If you found errors, fix them and return the 
corrected version.

Output only the corrected JSON. Do not add explanations.
\end{pybubblelistings}

\subsection*{B.5 Implementation Notes}

\begin{itemize}
\item All prompts instruct models to output only valid JSON without explanations or markdown formatting
\item Event-specific prompts (B.3) require one API call for event identification plus N additional calls for N identified events
\item Reflexion prompts (B.4) require two sequential API calls: initial extraction followed by verification
\item Zero-shot (B.1) and few-shot (B.2) prompts require only a single API call per sentence.
\end{itemize}

\end{document}